\documentclass[letterpaper,fleqn]{article} 
\usepackage{ist}
\usepackage{amsmath}
\usepackage{amssymb}
\usepackage[american]{babel}
\usepackage{microtype}
\usepackage{graphicx}
\usepackage{algorithm}
\usepackage{algorithmic}
\usepackage{siunitx}
\usepackage{subcaption}
\usepackage{float}
\usepackage{pdfpages}
\usepackage{tabularx}
\usepackage{multirow}
\usepackage{cite}

\DeclareMathOperator*{\argmax}{arg\,max}

\title{
Designing a Color Filter via Optimization of Vora-Value for Making a Camera more Colorimetric \\
}

\author{Yuteng Zhu, Graham D. Finlayson	\\
School of Computing Sciences, University of East Anglia, Norwich, UK \\}

\begin{document} 

\maketitle 
\thispagestyle{empty}
\begin{abstract}
The Luther condition states that if the spectral sensitivity responses of a camera are a linear transform from the color matching functions of the human visual system, the camera is colorimetric. Previous work proposed to solve for a filter which, when placed in front of a camera, results in sensitivities that best satisfy the Luther condition. 
By construction, the prior art solves for a filter for a given set of human visual sensitivities, e.g.\  the XYZ color matching functions or the cone response functions. However, depending on the target spectral sensitivity set, a different optimal filter is found.

This paper begins with the observation that the cone fundamentals, XYZ color matching functions or any linear combination thereof span the same 3-dimensional subspace. Thus, we set out to solve for a filter that makes the vector space spanned by the filtered camera sensitivities as similar as possible to the space spanned by human vision sensors. We argue that the Vora-Value is a suitable way to measure subspace similarity and we develop an optimization method for finding a filter that maximizes the Vora-Value measure.

Experiments demonstrate that our new optimization leads to filtered camera sensitivities which have a significantly higher Vora-Value compared with antecedent methods.

\end{abstract}

\section{1. \:Introduction} 

Color cameras are not colorimetric. That is, they do not satisfy the Luther condition~\cite{bibLuther}: their spectral sensitivities are not a linear transform from the CIE XYZ color matching functions (or equivalently any linear combination of the CMFs)~\cite{cameras}. Only cameras that satisfy the Luther condition can measure colors  in all conditions  which can then be corrected, without error, to colorimetric counterparts~\cite{exactHorn}.

\begin{figure}[t]
    \centering
    \includegraphics[width=0.2\textwidth]{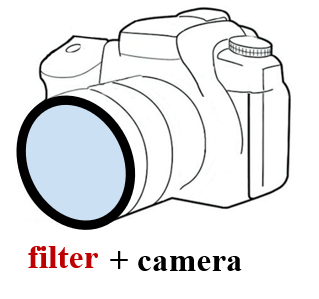}
    \caption{Illustration of the idea of placing a color filter in front of a camera for measurement.}
    \label{fig:filteredCamera}
\end{figure}

There are many papers in the literature address the colorimetric problem by proposing methods that design optimal filters to accurately recover tristimulus color responses~\cite{Trussell94,Trussel95,Wolski96,Vora97,Sharma97,Trussell97,Allebach2000,VoraInner}.
Recently, Finlayson \emph{et al.}~\cite{bibLutherFilter} developed a method for finding the best transmissive filter that, when placed in front of a camera (visualised in Figure~\ref{fig:filteredCamera}), results in the new effective sensitivities that best meet the Luther condition.
Curiously, however, if we change the target sensitivity functions - e.g.\ from XYZ CMFs to the cone fundamentals, themselves a linear transform apart~\cite{colorbibel}, the solved-for best filter changes.

\begin{figure}[t]
    \centering
    \includegraphics[width=0.475\textwidth]{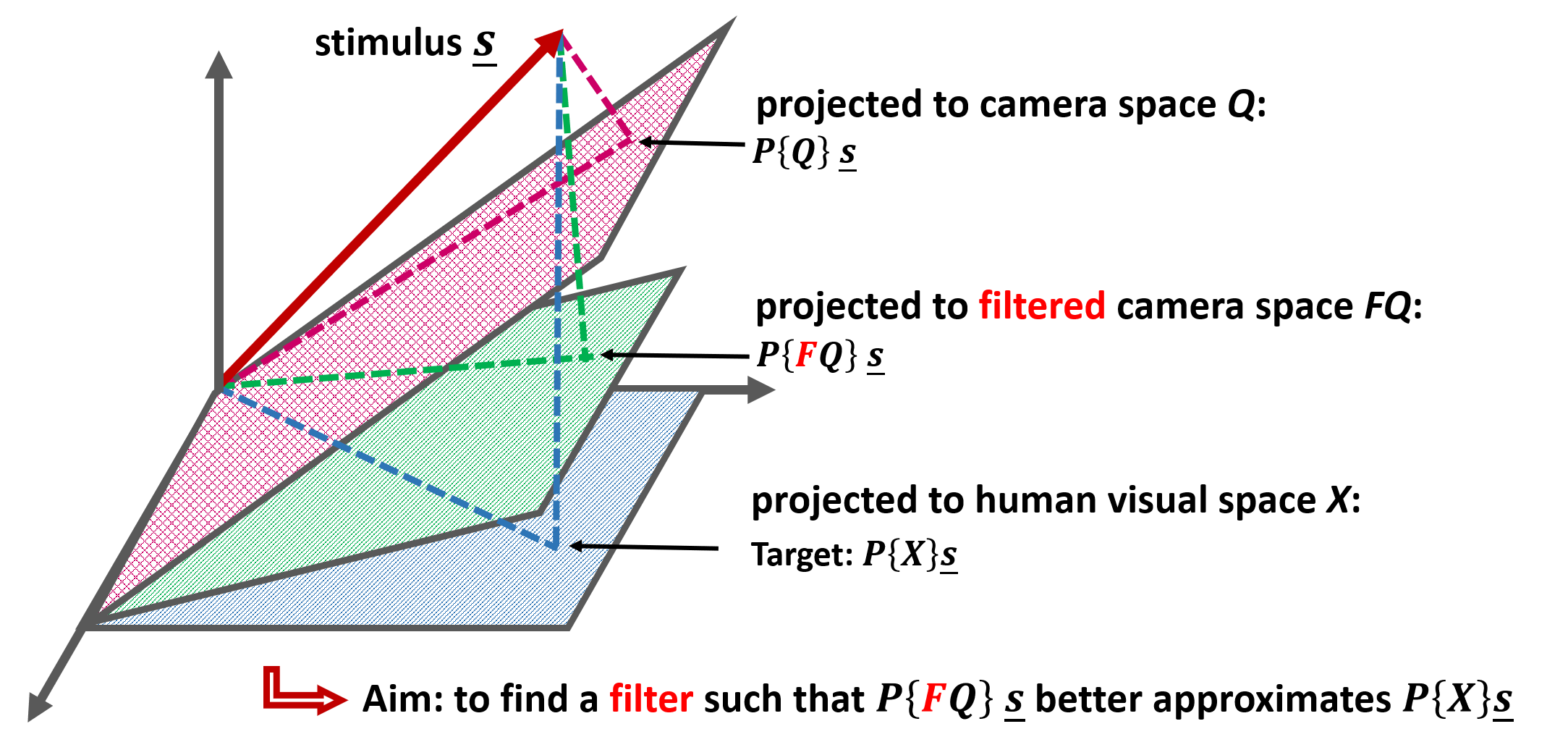}
    \caption{    A visualisation of the subspaces of camera space (in red), filtered camera space (in green), and the human visual space (in blue).}
    \label{fig:illustration}
\end{figure}

In this paper, we develop a method for finding a filter which makes the vector space spanned by the effective camera spectral sensitivities closest to all possible target sensitivity function sets for the Human Visual Space (HVS). We illustrate our approach in Figure~\ref{fig:illustration}. 
In blue, we show the vector space spanned by the sensitivity functions of the HVS. 
These sensitivities might be the cone fundamentals or equivalently the XYZ CMFs (we say equivalently because these sensitivity sets are linearly related and both sets span the same subspace). 
Similarly, the camera spectral  sensitivities - shown in red - span a different vector space.

When we place a filter in front of a camera, the new effective spectral sensitivities span a new subspace, shown in green in Figure~\ref{fig:illustration}. In this paper, we set forth a method for finding a filter 
which makes the filtered sensitivities spanning a subspace (the green vector space) as close as possible to the human visual space (the blue vector space).

Let's start to make some of this intuition more concrete. First, let us introduce a little notation. We denote respectively the spectral sensitivities of the camera and the human visual sensors as $\mathbf{Q}$ and  $\mathbf{X}$. For the purposes of this paper, we can think of these sensor sets as tables of discrete measurements. 
The columns of $\mathbf{Q}$ are the spectral sensitivities of the Red, Green and Blue sensors. Each row of $\mathbf{Q}$ is the three camera responses at discrete wavelengths. Usually, we measure sensitivities  from 400\,nm to 700\,nm 
(the typical wavelength range of visible spectrum) with a 10\,nm sampling distance, so the spectral sensitivity tables (or matrices) have 31 rows. $\mathbf{X}$ is similarly defined where columns could be the XYZ color matching functions or any function set thereof.

The blue and red planes in Figure~\ref{fig:illustration} - for the human visual system and camera sensitivities, respectively - visualise the subspaces spanned by $\mathbf{Q}$ and $\mathbf{X}$. 
By span we mean the set of all spectra we can generate by taking linear combinations of the columns of $\mathbf{Q}$ (or $\mathbf{X}$). We call a subspace because the columns of $\mathbf{Q}$ (or $\mathbf{X}$) span a 3-dimensional vector space contained in the 31-dimensional space. 
So, how do we evaluate the closeness (or otherwise) of two subspaces  spanned by the column vectors of $\mathbf{Q}$ and $\mathbf{X}$?

Usually, when we compare quantities, they are in {\it correspondence}. So far the ideas of vectors spanning a subspace has been quite abstract. The key to comparing  subspace is defining a method of placing spectra generated in one subspace in correspondence  to matching spectra in another. Let us  generate a 31-dimensional spectrum $\mathbf{\underline{s}}$. Returning to Figure~\ref{fig:illustration}, we denote the projection of $\mathbf{\underline{s}}$ onto $\mathbf{Q}$ as $P\{Q\}\mathbf{\underline{s}}$.  The corresponding projection onto $\mathbf{X}$ is denoted $P\{X\}\mathbf{\underline{s}}$. Finally, we denote the spectral sensitivities of a camera with a filter,  $\mathbf{F}$, placed in front as $\mathbf{FQ}$ and the corresponding projected spectrum as $P\{FQ\}\mathbf{\underline{s}}$.
For the stimulus $\mathbf{\underline{s}}$, now we  find the corresponding (projected) spectra, $\mathbf{\underline{s}_Q}$ and $\mathbf{\underline{s}_X}$ 
in the vectors spaces spanned by $\mathbf{Q}$ and $\mathbf{X}$ that are closest to $\mathbf{\underline{s}}$. 
And, given these corresponding spectra, we can calculate 
their distance: $||\mathbf{\underline{s}_Q}-\mathbf{\underline{s}_X}||_2$. By generating lots of random spectra - the values of which are independent and identically distributed -  drawn from 31-dimensional space, we can calculate the mean difference between their {\it projections} to the spaces spanned by $\mathbf{Q}$ and $\mathbf{X}$. This 
projected spectral error (after normalization) - calculated over all possible spectra - is our definition of subspace similarity.

This reasoning leads to a simple, closed-form, subspace similarity  formula that calculates a number in [0,1]. Intuitively, it is kind of a 
percentage measure of how similar two subspaces are. Actually, to this point our reason is (perhaps with a slightly different emphasis) a retelling of the derivation of the Vora-Value~\cite{bibVora}. The Vora-Value is a measure of how well a camera's RGBs can be (linearly) corrected to colorimetric counterparts (under the so-called Maximum Ignorance Assumption~\cite{MIA} which posits that all spectra are equally likely). Equivalently, it can be thought of as a percentage type measure of the similarity of two vector spaces, here the space spanned by two sensor sets (a camera and the HVS).

In this paper, we formulate an optimization expression using  the Vora-Value formula which we can maximize to find the {\it best} color filter to place in front of a camera
(i.e.\ the one that returns the best subspace similarity). The optimization method itself  is not closed-form but we show we can differentiate the objective function (with respect to the filter we seek) and then find the filter by a gradient ascent method~\cite{bibBoyd}. 

Experiments demonstrate that 
the Vora-Value score for many cameras is increased significantly by the optimized filters returned by our optimization. Further, the current method gives higher Vora-Values and consequently lower color measurement errors than when the filters are solved by the prior Luther-condition method~\cite{bibLutherFilter}.

In this paper, Section II reviews the definition of Vora-Value in detail and briefly introduces the gradient descent optimization method. We maximize the Vora-Value by a  gradient ascent approach in Section III. The experimental results and discussions are given in Section IV. The paper concludes in Section V.

\section{2. \:Background}
\vspace{0.1in}
\subsection{Vora-Value}

The Vora-Value of a camera sensor set $Q$ and the XYZ color matching functions $X$
is defined~\cite{VVmethoddesign} as
\begin{equation}
\begin{split}
      \nu (Q,X) &= \frac{trace(P\{Q\}P\{X\})}{trace(P\{X\})} 
\end{split}
\label{eq:vv_def}
\end{equation}
where $P\{Q\}$ and $P\{X\}$ denote the projection matrices of the camera sensitivity functions and the color matching functions, respectively and $trace()$ is the trace of a square matrix that sums up the elements along the diagonal of a matrix. In the equation, $trace(P\{X\})$ is a normalization term and so the Vora-Value is in the range of [0, 1]. For a trichromatic vision system where $X$ has three independent sensitivity functions, then $trace(P\{X\})=3$ (the trace of a projector is equal to the rank of the matrix).

The best case is when the subspace spanned by the camera sensitivities is the same as the  subspace spanned by the color matching functions then\ $P\{X\} = P\{Q\}$. Noting idempotency property of projectors $P\{X\}P\{X\}=P\{X\}$, the Vora-Value equals 1.  When the  camera sensitivity curves are  perpendicular to  - in the null space of -  the color matching functions, i.e.\ $P\{Q\}=I-P\{X\}$, the Vora-Value is 0. The Vora-Value can be thought of as a percentage 
measure of the similarity between two subspaces (see discussion in the Introduction) in general and of sets of spectral sensitivities in particular.

The projection of an $n\times m$ matrix $A$ with rank $m$ ($n>m$)  is defined as:
\begin{equation}
    P\{A\}=A(A^T A)^{-1}A^T
    \label{eq:proj}
\end{equation}
where the superscripts $^T$ and $^{-1}$ denote respectively the matrix transpose and inverse.
Substituting the explicit definition of a projector into Eq.~\ref{eq:vv_def} gives

\begin{equation}
\begin{split}
     \nu (Q,X) &= \frac{1}{3} trace({Q(Q^T Q)^{-1} Q^T X(X^T X)^{-1}X^T}).
\end{split}
\label{eq:vv_def_proj}
\end{equation}

Later we will show how we can find a filter that when placed in front of a camera results in effective camera sensitivities that yield a much higher Vora-Value ( compared to the unfiltered camera sensitivities). Our optimization is solved iteratively: we follow the gradient calculated from an optimization statement.

\subsection{Gradient Descent}
Let us illustrate the idea behind the gradient descent minimization.
Take the simple function  $f(x) = x^2$ that we would like to minimize. From the rules of calculus we can differentiate and find the zero for the 1st derivative and (using the 2nd derivative test) we can convince ourselves that the minimum of this function can be found at  $x = 0$.

\begin{figure}[t]
    \centering
    \includegraphics[width=0.3\textwidth]{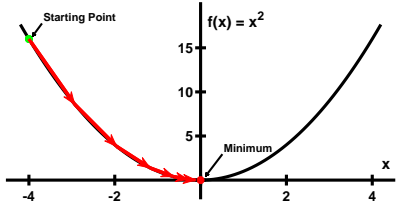}
    \vspace{0.05in}
    \caption{Illustration of gradient descent via a simple function.}
    \label{fig:GD_demo}
\end{figure}

Often however we cannot find the minimum (or minima) in closed form. This is where we use optimization techniques such as gradient descent.
The gradient descent process is illustrated in Figure~\ref{fig:GD_demo}. The arrows show how the function decreases when we use the gradient of the function to predict a lower value than where we currently are. By repeatedly following the gradient downwards we eventually find ourselves in a minimum. 

For an objective function with variables in n-dimension, 
the algorithm works in the same way except the gradient of the objective function becomes an $n$-dimensional vector, $\nabla f(\underline{x}) \in \mathbb{R}^n$.

In this paper we will be maximizing the function. It is the same idea but we follow the arrows `up' rather than down.

\section{3. \:Problem Statement and Algorithm}
\vspace{0.1in}
\subsection{Filter-modified Vora-Value}
When a filter is placed in front of a camera, it can be modeled as the multiplication of the filter spectral transmittance and the camera spectral sensitivities. On a per wavelength basis, 
let $ [r(\lambda), g(\lambda), b(\lambda)]$ and $f(\lambda)$ denote respectively the camera spectral sensitivities and the filter transmittance at a sampled wavelength. Then the new effective sensitivities of the filtered camera are equal to $[f(\lambda) r(\lambda), f(\lambda)g(\lambda), f(\lambda)b(\lambda)]$.

For our optimization it is useful to represent spectral quantities as vectors.  The filter vector $\underline{f}$ denotes an $n$-dimensional vector of transmittances across the visible spectrum. Since we typically sample the visible spectrum from 400nm to 700nm at every 10nm interval then $n$ is 31. Together with camera spectral sensitivity matrix $Q$, the new effective sensitivity responses after filtering can be represented as $FQ$ where the filter vector  is converted into a diagonal matrix $F = diag(\underline{f})$ for mathematical convenience, with its $i$th diagonal equals to the $i$th element of $\underline{f}$, i.e.\ $F_{ii} = f_i$. 

Thus the filter-modified Vora-Value for the effective `filter+camera' system can be written as
\begin{equation}
     \nu (FQ, X) = \frac{trace(P\{FQ\}P\{X\})}{trace(P\{X\})}
\label{eq:vv_fcam}
\end{equation}
Remind that $trace(P\{X\}) = 3$ for a trichromatic vision system.
Substituting the  projector matrices as Eq.~\ref{eq:proj}, we have
\begin{equation}
     \nu (FQ,X) =\frac{1}{3} trace(FQ(Q^T F^2Q)^{-1}Q^T F X(X^T X)^{-1}X^T).
\label{eq:vv_fcamexpand}
\end{equation}
Note in writing this equation we made us of the fact that  $F^T=F$ (diagonal matrices are symmetric).

\subsection{Constraining the Shape of the Filter}

We would like to control  the shape of the transmittance filter. Let us assume that a filter is a liner combination of a set of $k$ basis filters:  
\begin{equation}
     \underline{f}=B\underline{c} 
    \label{eq:filterBasis}
\end{equation}
where the columns of matrix $B$ denote the basis and $\underline{c}$ is a $k\times 1$ vector denoting the coefficients to each associated filter basis.

By judicious choice of basis, we can effectively bound the smoothness of the filter solution. Here we use the first $k$ terms in a Cosine Basis  expansion~\cite{cosine} -- which are smooth functions. 

\subsection{Bounding Filter Transmittance}
 Physically, the transmittance of a filter should be within the range of 0 and 100\% meaning fully absorptive and transmissive , respectively. However, if our algorithm finds a filter that transmits little light, it will have limited practical utility for our proposal of using the `filter+camera' system in accurate color measurement. 

In practice, we would like the filter transmittance in the desired bounds: ${f}_{min} \preceq \, \underline{f} \, \preceq {f}_{max} $. Here we use the symbol $\preceq$ to represent the vector inequality where every element in the vector satisfies the inequality. The scalars ${f}_{min}$ and ${f}_{max}$ denote the lower and upper thresholds on the desired transmittance of the filter. 

\subsection{Optimization Formulation and its Derivative}

The objective of our optimization is simply to find a filter $\underline{f}$ that maximizes the Vora-Value
\begin{equation}
   \argmax_{F} \; \nu (FQ,X)
    \label{eq:vv_max}
\end{equation}
where $Q$ and $X$ are $31 \times 3 $ matrices and $F$ is a $31\times 31$ diagonal matrix where the filter elements are the diagonal components as $F=diag(\underline{f})$. In the projector expanded form, the optimization is written as:

\begin{equation}
 \begin{split}
      \argmax_{F} \;  trace(FQ (Q^T F^2 Q )^{-1} Q^T F 
      X(X^T X)^{-1}X^T). 
  \label{eq:maxCoefs}  
 \end{split}   
\end{equation}
Strictly speaking, we should also divide by 3 (so the Vora-Value is between 0 and 1 but we ignore this step as a scalar does not affect the optimization).

The derivative of the objective function $\nu(FQ,X)$ with respect to the filter is captured by:
\begin{multline}
    \frac{\partial \nu(FQ,X) }{\partial F_{ii}} = \, [X(X^T X)^{-1}X^T FQ (Q^T F^2Q)^{-1}Q^T - \\
    FQ (Q^T F^2Q)^{-1}Q^T F X(X^T X)^{-1}X^T FQ (Q^T F^2Q)^{-1}Q^T ]_{ii}
    \label{eq:vv_Fii}
\end{multline}
Note because our filter is represented by a diagonal matrix: $F=diag(\underline{f})$, we are only interested in the partial derivatives along the diagonal $F$. 
Space does not permit a full recapitulation of the steps involved in differentiating $\nu$. Mainly we simply employ well known rules of matrix derivatives. We direct the reader to~\cite{bibmatrix} for a review.

Rewriting Eq.~\ref{eq:vv_Fii} using our projector matrix notation:
\begin{equation}
\frac{\partial \nu(FQ,X) }{\partial F_{ii}} =  [(I - P\{ FQ\})P\{X\}P\{ FQ\}F^{-1}]_{ii}
\label{eq:grad_F}
\end{equation}

Now, let us rewrite the derivative in terms of the underlying filter vector $\underline{f}$. First, remember that $F = diag(\underline{f})$. Here the $diag()$ function takes an $n$-component vector and places it along the diagonal of an $n\times n$ diagonal matrix. Let us now denote the inverse operator - the one that extracts the diagonal from a square matrix and places the result in a vector - as $vecd$. Clearly,
$vecd(diag(\underline{f}))=\underline{f}$. Please also bear in mind that $vecd()\neq vec()$. The latter is often used in the Kronecker Product extension to matrix algebra and maps an $n\times m$ matrix into an $nm\times 1$ vector~\cite{bibmatrix}. Here, $diag()$ is a forward operation turning a vector into a diagonal matrix and $vecd$ is the companion reverse operator extracting the diagonal.

We can write:
\begin{equation}
   \frac{\partial \nu(diag(\underline{f})Q,X) }{\partial \underline{f}} = vecd(  (I - P\{ FQ\})P\{X\}P\{ FQ\}F^{-1}) 
   \label{eq:grad_f}
\end{equation}
where the diagonal values of $(I - P\{ FQ\})P\{X\}P\{ FQ\}F^{-1}$ are taken to form the gradient function with respect to the filter vector $\underline{f}$. 

When $\underline{f} = B\underline{c}$, using the chain rule, the gradient function with respect to the coefficient vector $\underline{c}$ is calculated as:
\begin{equation}
 \frac{\partial \nu(diag(B\underline{c})Q,X) }{\partial \underline{c}} 
    = B^T  \frac{\partial \nu(diag(\underline{f})Q,X) }{\partial \underline{f}})
\end{equation}
or, equivalently in its explicit form as
\begin{multline}
     \frac{\partial \nu(diag(B\underline{c})Q,X) }{\partial \underline{c}} =  B^T vecd((I - P\{diag(B\underline{c})Q\})\; P\{X\} \\ P\{diag(B\underline{c})Q\} \; diag(B\underline{c})^{-1})   
\end{multline}

We make two final remarks. First, clearly the gradient has a very interesting structure: it is the diagonal of the product of 3 projection matrices multiplied by the inverse of the filter. Second, since we have an explicit gradient representation we are in control of our own maximization (see details in the next section). This is important as in future work we aim to investigate the optimization in more detail including the likelihood that there are potentially many local maxima and that these maxima could be closely located.

\subsection{Gradient Ascent }

We are maximizing the Vora-Value, so given the gradient function (calculated in the past section), we find the best filter by gradient ascent. Details are given in Algorithm 1. An important detail is the step size. The gradient vector of the function being maximized encodes the direction we should move to make the objective function larger. However, it does not teach directly how far we should move. If we add too much of the gradient we can step over the maximum and never reach the maximum point. Here the step size changes per iteration and is chosen by the backtracking line search algorithm~\cite{bibBoyd}.

\begin{algorithm}
	\caption{Gradient Ascent Algorithm}
	\label{algo_GD:f}
	\begin{algorithmic}[1]
		\STATE{$k=0, \quad \underline{f}^{0} =\underline{f}^{guess} $}
		\REPEAT
		\STATE{$k \gets k+1$}
		\STATE{Compute the Gradient: $\nabla \nu(\underline{f}^k)$}
    	\STATE{Choose the stepsize: $t ^{k}>0$}
		\STATE{Update $ \underline{f}^{k+1} = \underline{f}^k + t^{k} \nabla \nu(\underline{f}^k$)}
		\UNTIL{$ max(|\nabla \nu(\underline{f}^k)|) \leq \eta$}
\RETURN {$\underline{f}^{k+1}$}
	\end{algorithmic}
\end{algorithm}

\begin{figure}[!t]
    \centering
   \includegraphics[width=0.43\textwidth]{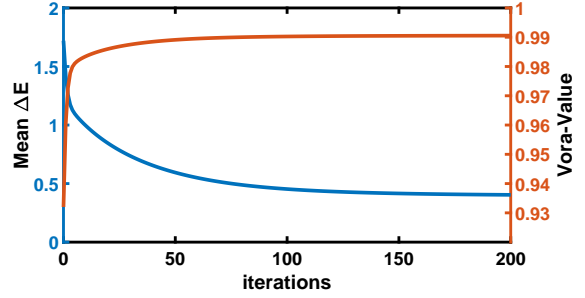}
    \vspace{0.1in}
    \caption{Algorithm convergence in terms of Vora-Value (right y-axis) and mean $\Delta E_{ab}^*$ with respect to iterations (left y-axis)}
    \label{fig:convergence}
\end{figure}

In Figure~\ref{fig:convergence}, we show how the gradient ascent algorithm converges for one camera (Canon 40D). On the x-axis we show the number of iterations of our algorithm. For each iteration we record the Vora-Value. We see here that our optimization converges quickly after just 100 iterations (see the red line associated with the right y-axis). 

Gradient method can also be carried out with inequality constraints. For example, in this paper we would like ${f}_{min} \preceq \, \underline{f} \, \preceq {f}_{max} $
where the interval $[f_{min},f_{max}]$ bounds the range of transmittance with respect to which we wish to find our filter. In the results shown in the next section we use the Projected Gradient Ascent Algorithm~\cite{PGD}, a simple variant of the gradient ascent shown above.

\section{4. \:Experiments and Results}
\vspace{0.1in}

\begin{figure}[!tb]
     \vspace{0.1in}
    \centering
     \includegraphics[width=0.45\textwidth]{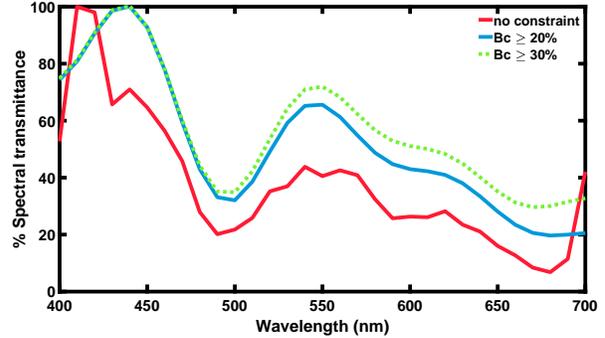}
     \vspace{0.1in}
    \caption{Spectral transmittance of Vora-Value optimized filters for Canon 40D camera. The filter in solid red line is solved with no constraint applied. Filters in solid blue and dotted green are modeled under 8-Cosine basis and constrained by minimum transmittance of 20\% and 30\%, respectively.}
    \vspace{0.1in}
    \label{fig:f4VV}
\end{figure}

Vora-Value optimized filters for a Canon 40D digital camera~\cite{bibsensor}  solved using Algorithm 1 are shown in Figure~\ref{fig:f4VV}.  In red is the filter found when no constraint (smoothness nor min max transmittances) is applied. In Blue, filters are described by the first 8 terms in a Cosine Basis expansion and constrained to be between 20\% and 100\% transmissive. Finally in dotted green we increase the minimum transmission to 30\%.

\begin{table}[!th]
\centering
\caption{Error statistics for Canon 40D camera}
\vspace{0.05in}
\renewcommand{\arraystretch}{1.3}
\setlength{\arrayrulewidth}{0.5mm}
\setlength\tabcolsep{4pt} 
\begin{tabular}{lp{1cm}lp{1cm}ccccc}
\hline \multirow{2}{*}{Method} & Vora-{ } 
& \multicolumn{5}{c}{color   errors $\Delta E_{ab}^{*}$}  \\ \cline{3-7} 
&Value  & mean  & median  & 95\% & 99\%  & max   \\ \hline
Baseline        & 0.932     & 1.72            & 1.03             & 5.12 & 12.94 & 28.39  \\ \hline
& \multicolumn{6}{c}{filter without   constraint} \\ 
Luther-filter & 0.986 & 0.44 & 0.22  & 1.48 & 3.19 &\bf 8.77   \\
Vora-filter  &\bf 0.991   &\bf 0.38 &\bf 0.20   &\bf 1.23 &\bf 2.97  & 9.89   \\\hline 
\\ [1pt] \hline
& \multicolumn{6}{c}{constrained filters $f=B\underline{c}$
}\\ 
& \multicolumn{6}{c}{condition I: 0.2  $\preceq \underline{f} \preceq 1$}  \\ 
Luther-filter &0.981 & 0.62 &0.38 &2.01 & 3.47 &\bf 9.52   \\
Data-driven  & 0.982	&\bf 0.45	&\bf 0.25	&\bf 1.41	&\bf 3.10	&10.63 \\
Vora-filter  & \bf 0.986 & 0.51  & 0.29 & 1.61 & 3.48  & 12.05  \\ 
& \multicolumn{6}{c}{condition II: 0.3 $\preceq \underline{f}  \preceq 1$} \\
Luther-filter &0.982  &0.69 & 0.41 &2.22 & 3.99 &\bf 12.69  \\
Data-driven  &0.983	&\bf 0.59	&\bf0.33	&\bf1.95	&\bf 3.76	&13.00 \\
Vora-filter &\bf 0.985   & 0.62  & 0.36  & 1.98 & 3.93  & 13.60 \\
\hline
\end{tabular}
\label{tab1}
\end{table}

In Table 1, we evaluate the optimized filters using our current method in terms of Vora-Value and compare the results with the prior art of Luther condition optimization~\cite{LIM2020} under the same constraint conditions. We also include the results of the unfiltered native camera sensor as baseline results. 
Using the current \textbf{Vora-filter}, Vora-Value of the effective camera sensitivities (to the reference CIE1931 XYZ color matching functions) increases to 0.991 from  0.932 for its unfiltered \textbf{native} sensor sensitivities and higher than 0.986 by the \textbf{Luther-filter}, see first three rows under the column of `\textbf{Vora-Value}' in Table 1. Remember that a higher Vora-Value indicates greater similarity of the subspaces spanned by the sensitivities of a camera and human visual system, therefore generally relates to more accurate color measurement~\cite{bibVora}.

We also evaluate the derived filters with respect to a  color measurement experiment. For a collection of 102 illuminants and 1995 refletance spectra~\cite{spectra}, we calculate the RGBs (for the native camera and the camera sensitivities after filtering) and ground-truth XYZs numerically.  The corresponding CIELAB color difference metric $\Delta E^*_{ab}$ statistics~\cite{colorbibel} are shown in columns 3-7 of Table 1. We can see that with a slight increase of Vora-Value, the current method reduces by about 10\% color errors (except max) compared to Luther-filter method. That is, for most color samples in the dataset, the current method outperforms the prior method.

We repeat this experiment where we optimize for shape-constrained smooth (linearly constructed by the first 8 cosine basis) and transmittance bounded filters. The corresponding Vora-Values and color measurement performance are shown in the second part of Table 1. 
Specifically for constrained filters, a data-driven method with multi-initialization conditions has been recently presented~\cite{EI2020}. The recent work designs a filter that is specifically optimal to the testing spectra dataset we are using here with the aim of minimizing the XYZ tristimulus error. Bear in mind that this `\textbf{Data-driven}' method uses 20,000 initialization filters seeding the algorithm for seeking the best filter refinement (and thus requires much greater computation). The corresponding $\Delta E^*_{ab}$ errors of `\textbf{Data-driven}' method are also included in Table~\ref{tab1} for reference. Save the max error, the other statistics are, perhaps unsurprisingly, leading in terms of $\Delta E^*_{ab}$.  Of course neither the Vora-filters (nor the Luther-filters) were derived to optimize $\Delta E^*_{ab}$.  However, the performance difference is relatively small (no more than about 10\% depending on the criterion). It is interesting that the Vora-Value method, a data-independent measure, performs so well in the color measurement even although it is not optimized for the testing dataset.

Returning to Figure~\ref{fig:convergence} where we recorded how the Vora-Value changed as we iteratively optimized for the best filter. Relative to our color measurement experiment, we can calculate the mean $\Delta E^*_{ab}$ for each per-iteration filter (multiplied by the Canon 40D spectral sensitivities). See how the average $\Delta E^*_{ab}$ also quickly diminishes after a small number of iterations.

For the prior art of Luther method~\cite{bibLutherFilter}, a filter is designed that best allows camera sensitivities to be linearly mapped to XYZ color matching functions. 
Interestingly, however, a better fit of the XYZ CMFs  does not predict lower $\Delta E^*_{ab}$ error in the color measurement test. Very speculatively, perhaps this is tentative evidence that the Vora-Value measure is a good predictor of the color measurement potential of a filter set (an idea proposed in the original paper~\cite{bibVora}).

\begin{figure}
    \centering
    \includegraphics[width=0.4\textwidth]{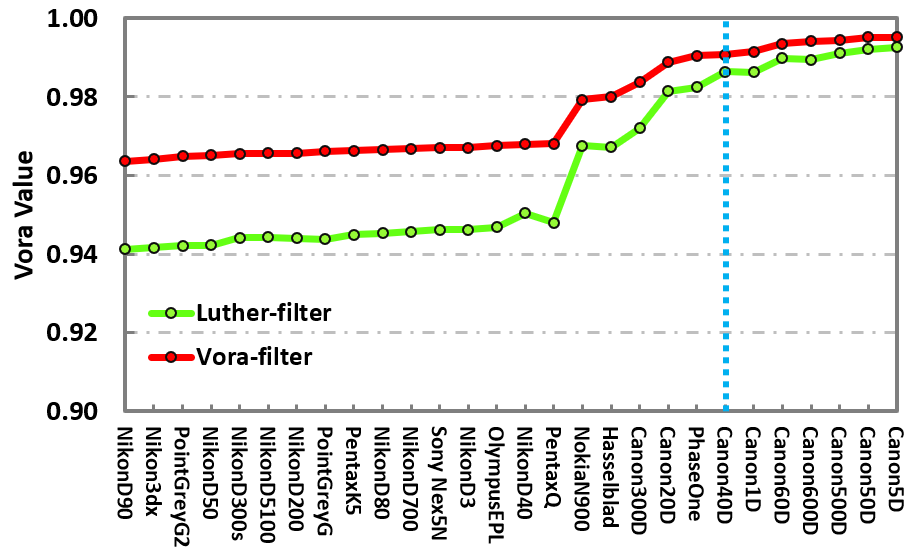}
    \caption{Evaluation of 28 cameras with optimized Luther-filters (in green) and Vora-filters (in red) in terms of Vora-Value.}
    \label{fig:VVfor28cams}
\end{figure}

Actually, the experiments we have presented so far, to some extent, hide the power of our method. In Figure~\ref{fig:VVfor28cams}, we plot the Vora-Value of 28 commercial cameras~\cite{bibsensor} with optimized Vora-filters and Luther-filters.  The dotted blue line points out where the Canon 40D camera we used for the experiments  is. We can see that the Canon 40D
can be well corrected with either the Luther method or (to a slightly higher number) by the optimization we developed in this paper. 
However, the figure shows that for many cameras, the difference between the Luther and Vora-Value methods are significant. Averaged over 28 cameras, Vora-filter improves the Vora-Value score to 0.976 from 0.918 for its unfiltered native camera sensors and higher than 0.961 (average correction) from Luther-filters.

\section{5. \:Conclusion}

In this paper, we set forth a filter design method such that a derived optimal filter, when placed in front of the camera, makes the camera much more colorimetric. The filter design problem is formulated as an optimization using the criterion of Vora-Value. 
Equivalently, by optimizing the Vora-Value we are moving the subspace spanned by the filtered camera spectral sensitivities closer to the subspace spanned by the Human Visual System sensors. 

Experiments validate our method. For all cameras tested we find the Vora-Value is significantly increased compared to antecedent methods. Color measurement experiments demonstrate that by optimizing the Vora-Value we also find the filtered sensitivities that support excellent color measurement.  When the derived filter is constrained to be smooth and have a bounded transmittance (e.g. at least 20\% of the light at any visible wavelength), we can still obtain the filtered camera sensitivities that have a high Vora-Value and, when used for color measurement, that support low color measurement errors.


\begin{biography}
Yuteng Zhu is currently a PhD candidate under the supervision of Prof. Finlayson in the Colour and Imaging Laboratory at the University of East Anglia, UK. She received her dual MSc degree from Zhejiang University, China, and \'Ecole Centrale de Marseille, France, in 2014.Her work has focused on filter design for accurate colour measurement by digital cameras.

Graham Finlayson is currently a Professor of Computing Sciences with the University of East Anglia. His research interests span color, physics-based computer vision, image processing, and the engineering required to embed technology in devices. He is a fellow of the Royal Photographic Society, the Society for Imaging Science and Technology, and the Institution for Engineering Technology.
\end{biography}

\end{document}